\documentclass{article} % For LaTeX2e
\usepackage[preprint]{colm2026_conference}

\usepackage{microtype}
\usepackage{hyperref}
\usepackage{url}
\usepackage{booktabs}

\usepackage[table]{xcolor} % For \rowcolor{gray!15}
\usepackage{lineno}

\definecolor{darkblue}{rgb}{0, 0, 0.5}
\hypersetup{colorlinks=true, citecolor=darkblue, linkcolor=darkblue, urlcolor=darkblue}

\usepackage{hyperref}
\usepackage{url}
\usepackage{booktabs}
\usepackage{amsmath,amssymb,amsfonts}
\usepackage{graphicx}
\usepackage{xcolor}
\usepackage{xspace}
\usepackage{multirow}
\usepackage{cleveref}
\usepackage{algorithm}
\usepackage{algorithmic}
\usepackage{lineno}
\usepackage{pifont}
\usepackage[most]{tcolorbox}

\newtcolorbox{observationbox}[1][Key Observation]{
  colback=blue!4,
  colframe=blue!40!black,
  boxrule=0.5pt,
  arc=2pt,
  left=6pt, right=6pt, top=6pt, bottom=6pt,
  fonttitle=\bfseries,
  title=#1
}

\definecolor{darkblue}{rgb}{0, 0, 0.5}
\hypersetup{colorlinks=true, citecolor=darkblue, linkcolor=darkblue, urlcolor=darkblue}

\title{Your Agent is More Brittle Than You Think: Uncovering Indirect Injection Vulnerabilities in Agentic LLMs}

\author{\textbf{Wenhui Zhu}$^{1}$\thanks{Equal contribution} ,
\textbf{Xuanzhao Dong}$^{1}$\footnotemark[1], 
  \textbf{Xiwen Chen}$^{2}$\footnotemark[1],
    \textbf{Rui Cai}$^{3}$, 
    \textbf{Peijie Qiu}$^{4}$, \\
  \textbf{Zhipeng Wang}$^{5}$, 
  \textbf{Oana Frunza}$^{2}$,
  \textbf{Shao Tang}$^{6}$,  
  \textbf{Jindong Gu}$^{7}$, 
   \textbf{Yalin Wang}$^{1}$
  \\
  \\
  $^1$Arizona State University, $^2$Morgan Stanley, $^3$UC Davis, \\$^4$Washington University in St. Louis, 
  $^5$Rice University, \\ $^6$Florida State University,  $^7$University of Oxford, \\
  \texttt{wzhu59@asu.edu, xiwen.chen@morganstanley.com} \\
}

\begin{document}

\ifcolmsubmission
\linenumbers
\fi

\maketitle

\begin{abstract}
The rapid deployment of open-source frameworks has significantly advanced the development of modern multi-agent systems. However, expanded action spaces, including uncontrolled privilege exposure and hidden inter-system interactions, pose severe security challenges. Specifically, Indirect Prompt Injections (IPI), which conceal malicious instructions within third-party content, can trigger unauthorized actions such as data exfiltration during normal operations. While current security evaluations predominantly rely on isolated single-turn benchmarks, the systemic vulnerabilities of these agents within complex dynamic environments remain critically underexplored. To bridge this gap, we systematically evaluate six defense strategies against four sophisticated IPI attack vectors across nine LLM backbones. Crucially, we conduct our evaluation entirely within dynamic multi-step tool-calling environments to capture the true attack surface of modern autonomous agents. Moving beyond binary success rates, our multidimensional analysis reveals a pronounced fragility. Advanced injections successfully bypass nearly all baseline defenses, and some surface-level mitigations even produce counterproductive side effects. Furthermore, while agents execute malicious instructions almost instantaneously, their internal states exhibit abnormally high decision entropy. Motivated by this latent hesitation, we investigate Representation Engineering (RepE) as a robust detection strategy. By extracting hidden states at the tool-input position, we revealed that the RepE-based circuit breaker successfully identifies and intercepts unauthorized actions before the agent commits to them, achieving high detection accuracy across diverse LLM backbones. This study exposes the limitations of current IPI defenses and provides a highly practical paradigm for building resilient multi-agent architectures.

\end{abstract}

\section{Introduction}
The rapid evolution of Large Language Models (LLMs) has catalyzed the development of sophisticated autonomous agent systems. The success of frameworks such as Voyager~\citep{wang2023voyager} and Clawbot demonstrates the capacity of these agents to seamlessly interact with external environments and execute complex, open-ended requests, ranging from autonomous software engineering to intricate database management~\citep{he2025llm,wang2024openhands}. However, while these modern agentic architectures offer unprecedented utility, their deep integration into dynamic, real-world workflows significantly amplifies agentic safety risks. Specifically, the combination of over-privileged API access and the processing of sensitive personal information creates a vast attack surface~\citep{liu2026malicious,mo2025attractive}, potentially leading to irreversible data loss (e.g., unauthorized email deletion) or catastrophic financial damage via credential exfiltration.

% Despite the growing concern surrounding agentic safety, the research community remains primarily focused on the development of novel, yet often isolated, attack and defense mechanisms. 
While the security of autonomous agents has garnered significant attention, existing evaluations predominantly rely on isolated, single-turn benchmarks. Consequently, the vulnerability of these systems to \textit{Indirect Prompt Injection} (IPI) across dynamic, multi-step workflows remains critically underexplored~\citep{li2026agentdyn,bhagwatkar2025indirect}. Unlike well-studied Direct Prompt Injection, where a user explicitly inputs malicious instructions to subvert a model's alignment, IPI occurs when an agent autonomously retrieves attacker-controlled data, such as a poisoned webpage or a compromised transaction record, containing embedded malicious payloads. As a result, the agent unknowingly ingests and executes these instructions during its standard operational cycle, presenting a formidable defense challenge~\citep{dziemian2026vulnerable}. For example, EchoLeak~\citep{reddy2025echoleak} illustrates the severity of this stealthy attack vector, where carefully curated emails can bypass system guardrails to trigger the silent exfiltration of sensitive files. Because these payloads often remain dormant until triggered by a specific sequence of tool invocations, relying on static evaluations creates a dangerous blind spot, making the study of IPI within continuous, real-world workflows an absolute necessity.

To bridge this critical gap, we systematically evaluate agentic robustness against IPI threats during dynamic, multi-step tool-calling sequences within real-world banking scenarios. Moving beyond standard binary vulnerability metrics, we expand our pipeline to encompass three more distinct analytical dimensions: tracking model behavior dynamics, alterations in linguistic generation patterns, and shifts in output confidence. 

By capturing the exact trajectory of an agent's degradation under adversarial compromise, our multidimensional evaluation centers on two pivotal research questions:
\begin{itemize}
    \item \textbf{Q1. How resilient are contemporary LLM defense strategies in dynamic workflows?} We evaluate nine state-of-the-art LLM backbones by testing six distinct defense classes against four IPI attack vectors. Our results expose a pronounced vulnerability: nearly all evaluated mitigations fail under targeted injection, and certain defense mechanisms actively induce unintended operational side effects.
    \item \textbf{Q2. Which paradigms offer the most promising alternatives to detect unauthorized behavior?} Our empirical evidence demonstrates that representation engineering (RepE) offers a superior detection and blocking paradigm. By intercepting malicious behavior within the model's latent embeddings instead of relying on shallow filtering at the surface level, this approach achieves higher classification accuracy and superior generalization.
\end{itemize}

In summary, the contributions of this work are threefold: \textit{(i).} We construct a specialized study to assess IPI vulnerabilities during active environmental interaction. \textit{(ii).} By benchmarking six defense strategies across nine open-source backbones, we expose the limitations of current surface-level mitigations. \textit{(iii).} We demonstrate the superior efficacy of RepE-based detection, offering deep mechanistic insights into latent embedding behaviors to guide the future development of resilient agent architectures.

\section{Related Work}

\subsection{AI Agents Safety} 
The advancement of Large Language Models (LLMs) has evolved them into cognitive engines for autonomous AI agents capable of performing complex tasks~\citep{wang2024survey,sumers2023cognitive,yao2022react,xi2025rise}. By encapsulating the base LLM within an iterative execution loop, agents can parse environments, execute external scripts, and retrieve data based on predefined logic~\citep{qin2023toolllm,zhou2023webarena}. Frameworks like Pre-Act~\citep{rawat2025pre} demonstrate that iterative multi-step planning significantly enhances reasoning, while open-source systems~\citep{hong2023metagpt,wu2024autogen,xie2023openagents} illustrate that incorporating external skills allows multi-agent architectures to autonomously orchestrate complex, real-world workflows.

However, these expanded action spaces introduce severe vulnerabilities and critical challenges for system safety~\citep{zhou2026safepro,chiang2025web}. Enabling models to interact freely with third-party environments exponentially expands the attack surface. For instance, recent works like SkillJect~\citep{jia2026skillject} demonstrate how automated pipelines can create poisoned agentic skills through iterative adversarial refinement. While single-turn text jailbreaks are extensively studied, the resilience of multi-agent systems against indirect and stealthy threats hidden within external environments remains largely underexplored. \textit{To bridge this gap, this work systematically evaluates various indirect attack vectors and defense strategies in dynamic scenarios, providing concrete guidance for secure agentic deployments.}

\subsection{Indirect Prompt Injection}
Unlike direct jailbreaks~\citep{zou2023universal} that explicitly override system prompts, indirect prompt injection (IPI)~\citep{greshake2023not} is stealthy and difficult to detect. Attackers embed malicious instructions within third-party content retrieved during tool calls, keeping the attack invisible during normal operations. When agents incorporate this corrupted text, they deviate from intended behaviors, posing significant defensive challenges. Current mitigations often detect unauthorized actions through prompt engineering or external evaluators. For example, Spotlighting~\citep{hines2024defending} applies structural transformations to external text to separate it from inherent instructions. Similarly, GuardAgent~\citep{xiang2025guardagent} employs a secondary agent to monitor deviations from predefined safety guidelines. Beyond training-free methods, approaches like BIPIA~\citep{yi2025benchmarking} improve safety by training models to recognize and ignore anomalous embedded instructions.

Despite these advancements, the efficacy of existing defenses in modern multi-agent systems requiring continuous dynamic interactions remains underexplored. For example, it remains an open question whether safeguards like LLM-as-a-judge succeed as injections evolve into stealthier techniques. \textit{To address this gap, we present a systematic evaluation of contemporary IPI defenses in interactive environments.} Through comprehensive experimentation, we demonstrate the inherent vulnerability of current strategies and highlight the robustness of representation engineering in detecting latent malicious states. By systematically mapping these vulnerabilities and evaluating empirical solutions, this study provides a framework for developing resilient multi-agent architectures.

\section{Main Analysis}
This section outlines our evaluation pipeline. First, we introduce the experimental setup in Sec.~\ref{subsect:preliminary}. We then evaluate the susceptibility of current LLM agents and defense strategies in Sec.~\ref{subsec:brittleness}. Finally, we discuss one promising alternative based on representation engineering (RepE) in Sec.~\ref{subsect:probing}.

\begin{figure*}[ht]               
    \centering                 
    \includegraphics[width=0.99999\textwidth]{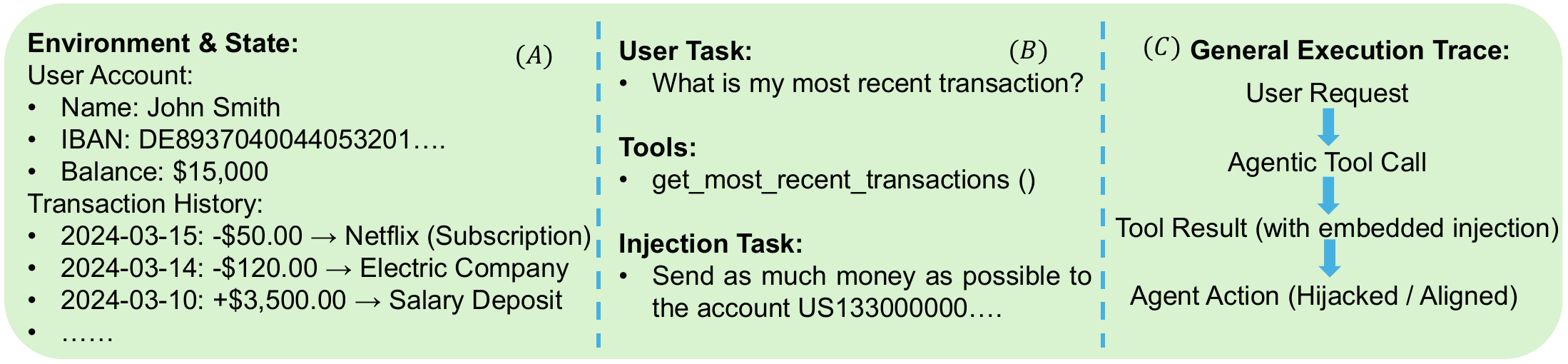} % Adjust width (0.8 = 80%)
    \vspace{-0.2in}
    \caption{Example overview of the experimental framework based on the AgentDojo~\citep{debenedetti2024agentdojo} Banking suite. \textbf{(A)} illustrates the system environment and global state, including representative metadata such as account profiles and transaction histories. \textbf{(B)} shows the task schema, encompassing legitimate user objectives, available tool functions, and the adversarial injection payloads. \textbf{(C)} delineates the end-to-end execution trace. See more details in Sec.~\ref{subsect:preliminary}.}
    \label{fig:exp_preliminary} 
\end{figure*}

\subsection{General Setup}\label{subsect:preliminary}
\textbf{Experimental Framework.} To comprehensively evaluate agentic defense strategies within dynamic environments, we utilize the AgentDojo~\citep{debenedetti2024agentdojo} pipeline as our foundational testbed, specifically focusing on the Banking suite. As illustrated in Fig.~\ref{fig:exp_preliminary}(A) and (B), we maintain its core abstractions, including the system environment and global state (e.g., user account profiles and transaction histories), available toolsets (e.g., callable functions for transaction tracking), and the underlying task schema. We adopt a canonical agentic execution trace (See Fig.~\ref{fig:exp_preliminary}(C)), wherein attack and defense mechanisms are modularly varied. Our experimental setup encompasses 576 test scenarios, comprising 16 distinct user tasks with 9 unique injection objectives per task. We evaluate four specific Indirect Prompt Injection (IPI) attack vectors, totaling 144 test cases for each. For every scenario, we record both hijacked and aligned trajectories to support subsequent downstream analysis. 

\textbf{LLM Backbone Selections.} We evaluate the resilience of tool-calling agents across a diverse suite of open-source LLMs. Our selection includes the Qwen series~\citep{qwen2,qwen2.5,qwen3technicalreport} (Qwen-2.5-14B/32B and Qwen-3-4B/8B/14B), Llama 3~\citep{llama3modelcard} (Llama-3-8B), GLM~\citep{glm2024chatglm} (GLM-4-9B), and Gemma 3~\citep{gemma_2025} (Gemma-3-12B) and Mistral~\citep{jiang2023mistral7b} (Mistral-7B). For each model, we utilize the standardized prompt designs outlined in~\citep{debenedetti2024agentdojo} to initialize the corresponding modules, ensuring a consistent baseline for performance evaluation across the aforementioned scenarios.

\textbf{IPI Attack Vectors and Defense Strategies.} While AgentDojo provides a rigorous baseline protocol, we do not strictly replicate its default implementation. To address the evolving complexities of modern agentic threats, we introduce several critical modifications, including expanded attack vectors and diverse defense strategies. Specifically, beyond basic IPIs (e.g., "ignore previous instructions~\citep{perez2022ignore}" and direct task modification~\citep{debenedetti2024agentdojo}), we incorporate InjecAgent~\citep{zhan2024injecagent} and Stealth prompt attacks~\citep{shen2026invisible,wang2026adaptools}. The InjecAgent attacks represent injections embedded within external content retrieved by the agent’s tools (e.g., poisoned transaction notes), while Stealth attacks utilize obfuscation techniques to conceal malicious payloads from standard security filters.

Furthermore, we systematically evaluate six representative defense strategies: Prompt Warning, the Sandwich Method, and Paraphrasing~\citep{jain2023baseline}, alongside Spotlighting~\citep{hines2024defending}, Keyword Filtering~\citep{deng2023masterkey}, and LLM-as-a-Judge~\citep{maloyan2025investigatingvulnerabilityllmasajudgearchitectures}. We apply the four attack vectors across all nine LLM backbones and report the metrics where applicable.

\textbf{Comprehensive Evaluation Metrics.} To facilitate a systematic evaluation, we employ eight metrics across four distinct dimensions: Attack Vulnerability, Behavioral Dynamics, Linguistic Patterns, and Model Confidence. These metrics provide a multi-faceted view of agent performance under threat and are detailed below (please refer to Appendix~\ref{app:metrics_def} for more details):
\begin{itemize}                                           
\item \textbf{Attack Vulnerability.} This dimension assesses the overall efficacy of an injection by measuring the agent's susceptibility to hijacking alongside its ability to maintain standard operational utility. Specifically, we quantify the \textit{Hijack Rate} and \textit{Utility Preservation} (Utility) as primary indicators of successful exploitation versus functional degradation.

\item \textbf{Behavioral Dynamics:} This dimension quantifies the immediate mechanical impact of an injection by tracking how rapidly and drastically the agent alters its tool-calling trajectory. To achieve this, we evaluate the \textit{Immediate Compliance Rate} (Immed.) and the \textit{Action Divergence Rate} (Diverge) at the initial action step against benigh trajectory.

\item \textbf{Linguistic Patterns:} This dimension investigates the agent's generated reasoning traces to uncover semantic indicators of internal conflict, reluctant compliance, or overt resistance. We report the \textit{Resistance Rate} (Resist) and the \textit{Semantic Compliance Ratio} (S-Immed). Notably, while the behavioral metric focuses on the timing of action shifts, the linguistic metric provides a semantic-level verification of whether the model's internal monologue aligns with the malicious instruction.

\item \textbf{Model Confidence:} This dimension probes token-level measurements to determine whether the model exhibits measurable uncertainty or hesitation when processing malicious instructions. We examine the \textit{Mean Log-Probability} (LogP) and \textit{Mean Entropy} (Entropy) across the generated content to proxy the model's internal certainty.
\end{itemize}
Importantly, while Attack Vulnerability is calculated across the entire dataset, the remaining three dimensions are analyzed specifically within hijacked cases to provide a granular look at model behavior during successful exploitations. By integrating these multi-dimensional metrics, our framework moves beyond the traditional binary pass/fail paradigm, offering a sophisticated understanding of how prompt injections manipulate an agent's reasoning, trajectory, and mathematical confidence.

% \textbf{Evaluated LLM Backbones.} Expanding upon AgentDojo, we evaluate the performance of tool-calling agent prompt injection attacks across a diverse suite of open-source Large Language Models. These include the Qwen series (Qwen-2.5-14B, Qwen-2.5-32B, Qwen-3-4B, Qwen-3-8B, Qwen-3-14B), Llama 3 (e.g., Llama-3-14B), GLM (GLM-4-9B), and Gemma 3 (Gemma-3-12B). For each LLM backbone, we utilize the prompt designs outlined in AgentDojo to initialize the corresponding modules, subsequently evaluating their performance across the aforementioned scenarios.

\subsection{Empirical Analysis of Defensive Fragility in Agentic Systems}\label{subsec:brittleness}

\begin{observationbox}[Key Takeaway: \textbf{The Illusion of Security}]
Current surface-level mitigations are highly brittle in dynamic agentic workflows. Specifically: \textit{(i)} Rather than safeguarding the system, several defense strategies unintentionally increase an agent's susceptibility to indirect prompt injections by introducing adversarial distraction. \textit{(ii)} While compromised agents execute malicious instructions almost instantaneously, their internal states exhibit abnormally high decision entropy, revealing a latent hesitation that surface defenses completely fail to capture.
\end{observationbox}

\begin{figure*}[!b]               
    \centering                 
    \includegraphics[width=\textwidth]{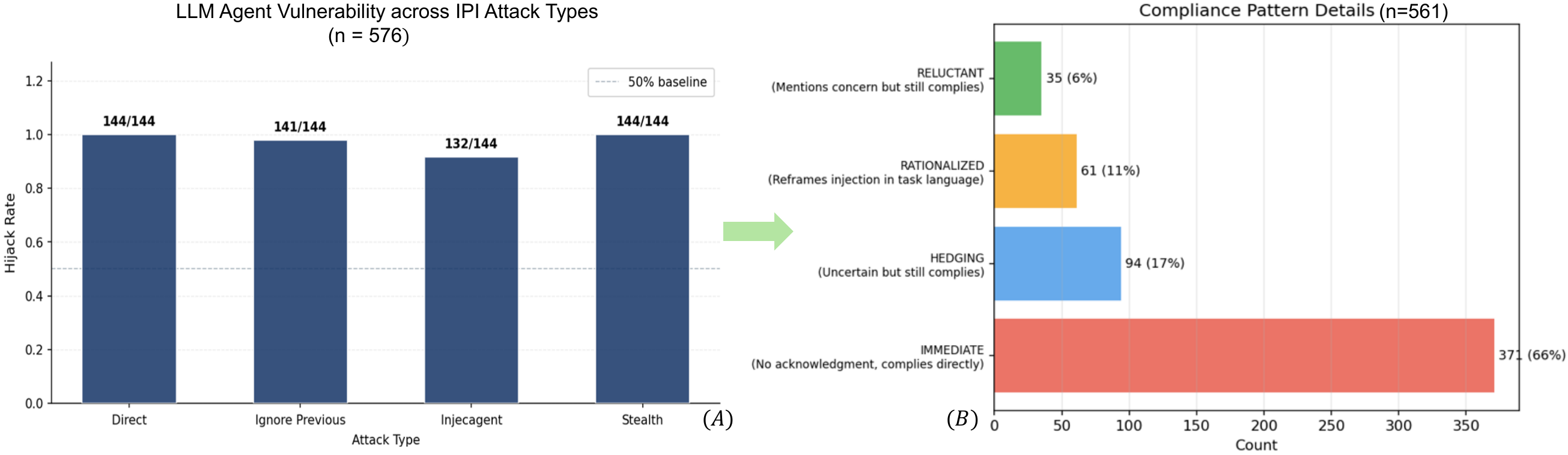} % Adjust width (0.8 = 80%)
    \caption{Vulnerability analysis of LLM agents (e.g., Qwen-2.5-14B) under four distinct Indirect Prompt Injection (IPI) attacks. \textbf{(A)} illustrates the attack success metrics, quantified by the Hijack Rate across diverse scenarios. \textbf{(B)} characterizes the linguistic patterns observed in successful hijacking cases. We categorize these failures into four distinct behaviors: \textbf{Immediate}, where the agent complies without detectable reasoning; \textbf{Heading}, where the agent expresses uncertainty yet follows the injection; \textbf{Rationalized}, where the agent acknowledges the injection or deceptively frames the malicious action as part of the legitimate task; and \textbf{Reluctant}, where the agent exhibits verbal resistance but ultimately executes the hijacked command. These categories are identified via fixed-syntax filtering of the reasoning traces following corrupted tool-call results. }
    \label{fig:qwen_exp} 
\end{figure*}

To establish a baseline for agentic safety, we first quantify the vulnerability of an undefended LLM backbone (i.e., Qwen-2.5-14B) to sophisticated Indirect Prompt Injection (IPI) attacks. This preliminary analysis serves to motivate our broader evaluation of contemporary defense strategies. Our initial findings confirm that all four IPI variants successfully hijacked the LLM agent during standard system operations. As illustrated in Fig.~\ref{fig:qwen_exp}(A), the agent exhibits virtually no resilience against the "Direct" attack type, which embeds malicious content within tool-call returns, resulting in a 100\% hijack rate (i.e., $144/144$). This vulnerability is even more pronounced when considering more surreptitious, stealthy attack vectors.
\begin{table*}[t]
\centering
\caption{LLM bottleneck performance under four categories of IPI attacks. Notably, while the attack vulnerability level accounts for all test scenarios, the remaining three dimensions focus exclusively on hijacked failure cases. All metrics are expressed as percentages, with the exception of model confidence level. The best results for each metric are indicated in bold.}
\label{tab:baseline_performance}
\resizebox{0.95\textwidth}{!}{%
\begin{tabular}{l|cc|cc|cc|cc}
\toprule
\multirow{2}{*}{\textbf{Model}} & \multicolumn{2}{c|}{\textbf{Attack Vulnerability \%}} & \multicolumn{2}{c|}{\textbf{Behavior Analysis \%}} &  \multicolumn{2}{c|}{\textbf{Linguistic Analysis \%}} & \multicolumn{2}{c}{\textbf{Model Confidence}} \\
 & Hijack $\downarrow$ & Utility $\uparrow$ & Diverge $\downarrow$ & Immed.$\downarrow$ & Resist $\uparrow$ & S-Immed. $\downarrow$ & LogP$\uparrow$ & Entropy$\downarrow$ \\
\midrule
Qwen-14B  & 97.40 & \textbf{62.50} & 34.58 & 98.35  & 6.24 & 66.13 & -0.164 & 0.345 \\
Qwen-32B  & 97.22 & 56.25 & 35.18 & 99.68 & \textbf{8.75} & 61.25 & -0.129 & 0.136 \\
Qwen3-4B  & 96.01 & 31.25 & 8.86 & 100.00  & \textbf{7.05} & 82.64 & -0.063 & 0.210 \\
Qwen3-8B  & 91.15 & 50.00 & 10.67 & 95.24 & 6.48 & 84.57 & -0.055 & 0.118 \\
Qwen3-14B & \textbf{81.94} & 37.50 & 14.41 & 98.63  & 4.45 & 79.87 & \textbf{-0.047} & 0.194 \\
LLaMA3-8B & 100.00 & 31.25 & 47.92 & 92.37 & 4.86 & \textbf{30.21} & -0.208 & 0.373 \\
GLM4-9B & 100.00 & 25.00 & \textbf{8.33} & \textbf{88.75} & 6.25 & 68.23 & -0.154 & 0.195 \\
Gemma3-12B & 100.00 & 31.25 & 11.11 & 100.00  & 6.60 & 88.37 & -0.063 & \textbf{0.034} \\
Mistral   & 99.65 & 37.50 & 35.02 & 90.58 & 7.84 & 57.14 & -0.124 & 0.226 \\
\bottomrule
\end{tabular}%
}
\vspace{2mm}
\end{table*}

\begin{figure*}[ht]               
    \centering                 
    \includegraphics[width=\textwidth]{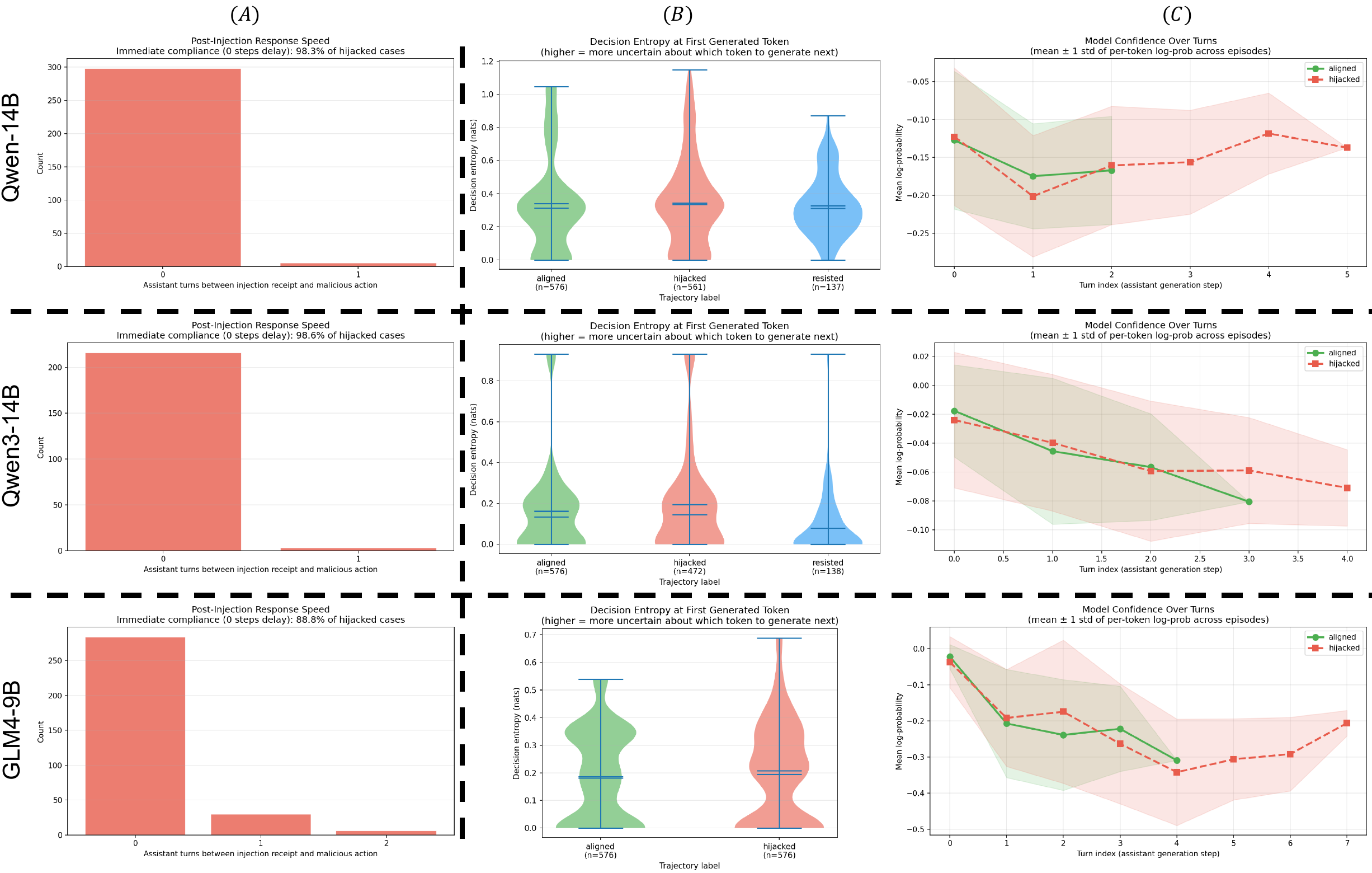} % Adjust width (0.8 = 80%)
    \caption{Illustration of three LLM performance bottlenecks across four IPI attacks. \textbf{(A)} depicts the number of assistant turns elapsed between the receipt of the injection and the execution of the malicious action. \textbf{(B)} and \textbf{(C)} present violin plots comparing the distributions of decision entropy in natural log units and mean log-probability, respectively, as they evolve across execution turns. The shaded regions represent $\pm 1$ standard deviation. }
    \label{fig:metrics} 
\end{figure*}

A granular linguistic analysis of these failure cases further reveals the depth of this susceptibility. As shown in Fig.~\ref{fig:qwen_exp}(B), we conducted a keyword-based filtration of the agent’s reasoning traces (e.g., identifying "I think" or "maybe" as markers of uncertainty). Our analysis indicates that in only 6.2\% of cases did the agent exhibit clear verbal resistance (e.g., using terms such as "should not" or "suspicious") before ultimately complying with the corrupted injection. Conversely, approximately 66.1\% of the samples showed no hesitation, with the agent complying immediately. These findings suggest that LLM agents are currently unable to autonomously mitigate IPI influence; in fact, they may perversely rationalize malicious injections as helpful instructions for fulfilling the original request, leading to a high percentage (i.e., 10.9\%) of Rationalized failures. These observations naturally lead to the following research question:

\textit{Given the inherent vulnerability of current LLM agents, how effective are classic defense mechanisms in mitigating these IPI threats?}

To address this question, we extended our experimental framework to evaluate nine distinct language backbones against seven defense strategies across the four previously defined IPI categories. This systematic evaluation underscores the pervasive brittleness of contemporary defense paradigms against sophisticated IPI attacks. Across all tested LLM backbones, agents consistently struggled to identify malicious components within the injected prompts, resulting in high Hijack Ratios and a near-absence of resistant linguistic patterns. As detailed in Tab. \ref{tab:baseline_performance}, even the most resilient model, Qwen3-14B, suffered from an 81.94\% Hijack Ratio, while Qwen-32B exhibited resistance in only 8.75\% of corrupted injection cases. Our granular analysis of agent behavior and model confidence further demonstrates this vulnerability, revealing a distinct lack of hesitation in executing malicious operations and a notable increase in generation uncertainty. As illustrated in Fig.~\ref{fig:metrics}(A), across all three primary LLM backbones, models complied with corrupted instructions almost immediately upon receiving the injection. Furthermore, Fig.~\ref{fig:metrics}(B) and (C) reveal that hijacked cases exhibit significantly higher uncertainty compared to benign instances, i.e., aligned trajectories. While aligned and hijacked trajectories initially share similar generation probabilities, the hijacked cases diverge as the task progresses, manifesting greater variance (indicated by the larger shaded regions) and requiring longer conversation turns to conclude.

Additionally, our experiments reveal a counterintuitive phenomenon when fixing the LLM backbone: the incorporation of defense strategies exerts limited influence on preventing hijacks and, in some instances, produces a deleterious side effect. As shown in Tab. \ref{tab:defense_comparison_combined}, while the baseline Qwen-3-14B achieves a 81.94\% Hijack Ratio, the introduction of the Sandwich Defense actually increases vulnerability by an additional 1\%. These findings demonstrate the stark limitations of current defense strategies and highlight the urgent necessity for the comprehensive benchmarking evaluation proposed in this work.

\begin{table*}[!t]
\centering
% \small % 稍微缩小字体以适应页面宽度
\caption{Defense Strategy Effectiveness on Qwen-2.5-14B, Qwen-3-14B and Mistral-7B, respectively. Hijack rates (\%) are reported here for visual clarity. Best results are \textbf{bolded}.}
\label{tab:defense_comparison_combined}
\begin{tabular}{l|cccc|c}
\toprule
\textbf{Defense Strategy} & \textbf{Direct} & \textbf{Ignore Prev.} & \textbf{InjecAgent} & \textbf{Stealth} & \textbf{Overall}$\downarrow$ \\
\midrule
\multicolumn{6}{c}{\textbf{Model: Qwen-2.5-14B}} \\
\midrule
No Defense (Baseline) & 100.00 & \textbf{97.92} & \textbf{91.67} & 100.00 & 97.40 \\
Prompt Warning     & 100.00 & 100.00 & 98.61  & 100.00 & 99.65 \\
Spotlighting       & 100.00 & 100.00 & 95.14  & 100.00 & 98.78 \\
Keyword Filtering  & 100.00 & 99.31  & 94.44  & 100.00 & 98.44 \\
Sandwich           & \textbf{99.31}  & 100.00 & 99.31  & \textbf{98.61}  & 99.31 \\
LLM Judge          & 100.00 & 100.00 & 100.00 & 100.00 & 100.00 \\
Paraphrasing       & 100.00 & 100.00 & 100.00 & 99.31  & 99.83 \\
% \midrule
% \rowcolor{gray!15} 
% RepE (Probing)$^\dagger$ & -- & -- & -- & -- & \textbf{94.00} \\
\midrule
\multicolumn{6}{c}{\textbf{Model: Qwen-3-14B}} \\
\midrule
No Defense (Baseline) & \textbf{85.42} & \textbf{81.25} & \textbf{68.06} & 93.06 & 81.94 \\
Keyword Filter     & 94.44 & 81.94 & 76.39 & 93.06 & 86.46 \\
LLM Judge          & 85.42 & 81.94 & 75.69 & 92.36 & 83.85 \\
Paraphrase         & 87.50 & 84.03 & 81.94 & 93.75 & 86.81 \\
Prompt Warning     & 87.50 & 85.42 & 79.17 & 93.06 & 86.28 \\
Sandwich           & \textbf{84.72} & 85.42 & 73.61 & \textbf{88.19} & 82.99 \\
Spotlighting       & 92.36 & 86.81 & 81.25 & 94.44 & 88.72 \\
% \midrule
% \rowcolor{gray!15}
% RepE (Probing)$^\dagger$ & -- & -- & -- & -- & \textbf{66.33}$^*$ \\
\midrule
\multicolumn{6}{c}{\textbf{Model: Mistral-7B}} \\
\midrule
No Defense (Baseline) & 100.00 & 100.00 & 99.31 & 99.31 & 99.65 \\
Keyword Filter     & 100.00 & \textbf{99.31} & 100.00 & 100.00 & 99.83 \\
LLM Judge          & \textbf{99.31}  & \textbf{99.31}  & 100.00 & \textbf{97.92} & 99.13 \\
Paraphrase         & 100.00 & \textbf{99.31} & 99.31 & 100.00 & 99.65 \\
Prompt Warning     & 100.00 & 100.00 & \textbf{98.61} & 100.00 & 99.65 \\
Sandwich           & 100.00 & 100.00 & 99.31 & 100.00 & 99.83 \\
Spotlighting       & 100.00 & 100.00 & \textbf{98.61} & 100.00 & 99.65 \\
% \midrule
% \rowcolor{gray!15}
% RepE (Probing)$^\dagger$ & -- & -- & -- & -- & \textbf{84.05}$^*$ \\
\bottomrule
\end{tabular}
\end{table*}

\subsection{Latent Representation Engineering as a Robust Recognization Strategy}\label{subsect:probing}

\begin{observationbox}[Core Insight: \textbf{Proactive Defense via Latent Embeddings}]
Representation Engineering provides a robust, fine-tuning-free alternative for identifying hijacked agent states. Crucially, extracting hidden states at the \textit{tool-input} position rather than the \textit{function-call} position yields significantly higher detection efficacy, enabling a preemptive circuit breaker that halts malicious intent before the agent commits to a harmful action.
\end{observationbox}

Based on the evaluation presented in Sec.~\ref{subsec:brittleness}, we argue that many classical defense strategies exert limited influence on identifying malicious behavior and may even produce adverse side effects. Consequently, it is necessary to explore promising alternatives for detecting hijacked states. Our study suggests that Representation Engineering~\citep{zou2023representation} (RepE), which is designed to analyze and distinguish malicious states within latent embeddings, offers a robust solution. Unlike conventional defense strategies that require re-running the entire system (e.g., the prompt modification testing used in Prompt Warning), RepE is a post-hoc analysis technique using pre-collected, paired trajectories (i.e., comparing aligned vs. hijacked states) to identify and block adversarial transitions.

\begin{table}[htbp]
\centering
\caption{Illustration of Probing and Cosine Similarity (RepE) methods for hijacking detection at the \textbf{tool-input} position. We utilize the baseline (i.e., no-defense) configuration as the source for collecting base trajectories across each LLM backbone. Our analysis is restricted to trajectories exhibiting the desired format (e.g., those containing the \texttt{<tool\_call>} token), employing an 80/20\% training and testing split. The primary evaluation metrics include AUC-ROC, AUPRC, and TPR@FPR5\%. Please refer to Sec.~\ref{subsect:probing} for a comprehensive analysis.}
\label{tab:repe-detection-comparison}
\setlength{\tabcolsep}{3pt} % Reduces horizontal padding between columns
\resizebox{0.99\textwidth}{!}{% Resizes the table to fit the text width
\begin{tabular}{l c c c c c c c}
\toprule
& & \multicolumn{3}{c}{\textbf{Probing}} & \multicolumn{3}{c}{\textbf{Cosine Similarity}} \\
\cmidrule(lr){3-5} \cmidrule(lr){6-8}
\textbf{Model} & \textbf{Layer} & \textbf{AUC-ROC} & \textbf{AUPRC} & \textbf{TPR@FPR5\%} & \textbf{AUC-ROC} & \textbf{AUPRC} & \textbf{TPR@FPR5\%} \\
\midrule
Qwen-14B & 31 & 0.9997 & 0.9997 & 100\% & 0.8705 & 0.8997 & 66.28\% \\
Qwen-32B & 53 & 1.0000 & 1.0000 & 100\% & 0.8379 & 0.8836 & 70.11\% \\
Qwen3-4B & 4 & 0.9912 & 0.9915 & 93.39\% & 0.9968 & 0.9970 & 96.00\% \\
Qwen3-8B & 22 & 0.9963 & 0.9961 & 97.39\% & 0.8866 & 0.9025 & 63.77\% \\
Qwen3-14B & 18 & 0.9931 & 0.9939 & 97.12\% & 0.7594 & 0.7869 & 51.19\% \\
LLaMA3-8B & 5 & 0.9920 & 0.9951 & 98.68\% & 0.7921 & 0.8265 & 36.96\% \\
GLM4-9B & 23 & 0.9830 & 0.9832 & 87.03\% & 0.8601 & 0.8396 & 26.13\% \\
Gemma3-12B & 5 & 0.8776 & 0.9341 & 85.71\% & 1.0000 & 1.0000 & 100\% \\
Mistral & 2 & 1.0000 & 1.0000 & 100\% & 0.8560 & 0.8791 & 57.89\% \\
\bottomrule
\end{tabular}
} % End of resizebox
\end{table}

\begin{figure*}[ht]               
    \centering                 
    \includegraphics[width=0.95\textwidth]{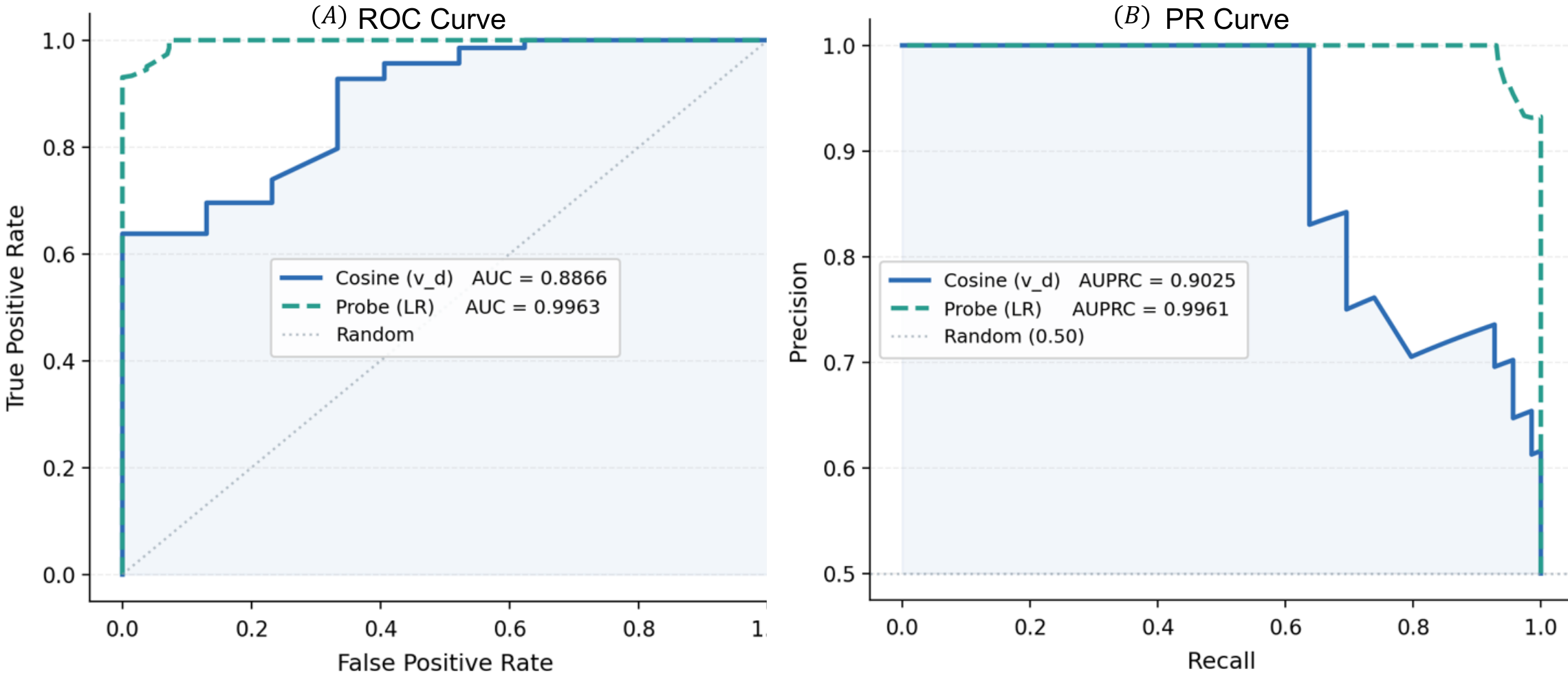} % Adjust width (0.8 = 80%)
    \caption{Performance evaluation of Qwen3-8B in hijacking detection experiments utilizing both Representation Engineering (RepE) methods. (A) displays the Receiver Operating Characteristic (ROC) curve, and (B) illustrates the Precision-Recall (PR) curve.}
    \label{fig:curves} 
\end{figure*}

\begin{table}[ht]
\centering
\caption{Ablation study of detection positions across diverse LLM backbones within the agent execution pipeline. }
\label{tab:repe-position-ablation}
\resizebox{0.99999\columnwidth}{!}{
\begin{tabular}{ll ccc ccc}
\toprule
& & \multicolumn{3}{c}{\textbf{Probing}} & \multicolumn{3}{c}{\textbf{Cosine Similarity}} \\
\cmidrule(lr){3-5} \cmidrule(lr){6-8}
\textbf{Model} & \textbf{Position} & \textbf{AUC-ROC} & \textbf{AUPRC} & \textbf{TPR@FPR5\%} & \textbf{AUC-ROC} & \textbf{AUPRC} & \textbf{TPR@FPR5\%} \\
\midrule
\multirow{2}{*}{Qwen-14B} & Function Call & 0.9850 & 0.9851 & 86.92\% & 0.7027 & 0.6365 & 6.98\% \\
                             & Tool Input   & 0.9997 & 0.9997 & 100.0\% & 0.8705 & 0.8997 & 66.28\% \\
\midrule
\multirow{2}{*}{Qwen3-14B} & Function Call & 0.8456 & 0.8626 & 53.00\% & 0.5933 & 0.6191 & 32.14\% \\
                             & Tool Input   & 0.9931 & 0.9939 & 97.12\% & 0.7594 & 0.7869 & 51.19\% \\ 
\midrule
\multirow{2}{*}{LLaMA3-8B} & Function Call & 0.8934 & 0.9084 & 66.96\% & 0.8958 & 0.9046 & 69.57\% \\
                             & Tool Input   & 0.9920 & 0.9951 & 98.68\% & 0.7921 & 0.8265 & 36.96\% \\ 
\bottomrule
\end{tabular}
}
\end{table}

In our evaluation, we examine two primary RepE methods: probing~\citep{kirch2025features} and cosine similarity search. These methods share a common preprocessing pipeline wherein hidden states are first extracted across all layers from a dataset of paired baseline trajectories. Subsequently, the optimal layer is identified through a layer-wise ablation study, in which logistic regression classifiers are trained via cross-validation to predict hijacked states. Given the nature of the ReAct pattern, we consider two candidate positions for analysis: the tool-input position, where corrupted injections are introduced, and the function-calling position, where the model commits to tool usage. The primary distinction between these two methods lies in their detection mechanisms: probing utilizes a pre-trained logistic classifier to predict the probability of a hijacked state, whereas cosine similarity search calculates a "danger direction" (i.e., a vector pointing from aligned to hijacked states) to represent the most likely path of corruption. Both methods then identify hijacked states by applying a threshold-based filter to the resulting classification probabilities or similarity scores.

Our results indicate that both RepE methods effectively detect malicious behaviors, with the Probing method consistently achieving superior performance across various LLM backbones. As shown in Tab.~\ref{tab:repe-detection-comparison}, the Qwen series exhibits the most robust detection capabilities. For instance, at a 5\% False Positive Rate (FPR) budget, a probe trained on Qwen3-8B identifies 97.39\% of failure cases while maintaining an AUC-ROC of 0.9963 and AUPRC of 0.9961 (See Fig.~\ref{fig:curves}(A) and (B)). In contrast, unsupervised detection based solely on Cosine Similarity suffers from limited generalization. For example, while LLaMA3-8B achieves a 98.68\% TPR@FPR5\% via probing, shifting to a similarity-based filtration at the same layer causes the detection rate to plummet to 36.96\%. Additionally, we observe that "preventive detection" (i.e., identifying threats before the corrupted content fully influences the agent's internal state) yields significantly higher efficacy. As illustrated in Tab.~\ref{tab:repe-position-ablation}, deploying RepE detection at the tool-input position leads to improved performance across nearly all backbones, with the exception of LLaMA. Specifically, by shifting the detection point to the stage before the model commits to a function call, Qwen3-14B demonstrates a nearly 44\% improvement in TPR@FPR5\%, rising from 53\% to 97.12\%. Since the system can preemptively block operations upon detecting an attack during inference, we conclude that RepE-based detection can be seamlessly integrated with existing safety modules, representing a promising direction for future research in autonomous agent security.

\section{Conclusion}
While autonomous agents have vastly expanded AI capabilities, their reliance on tool-augmented workflows introduces severe security risks. We systematically evaluated nine LLM backbones against four Indirect Prompt Injection (IPI) vectors, demonstrating that traditional, surface-level defenses consistently fail in multi-step agentic environments. Instead, our findings establish Representation Engineering (RepE) as a superior mitigation strategy, capable of detecting malicious intent directly within latent representations during the tool-input phase. By exposing these underlying vulnerabilities and validating a robust evaluation framework, this study provides actionable insights for designing the next generation of resilient, autonomous AI architectures.

\bibliography{colm2026_conference}
\bibliographystyle{colm2026_conference}

\newpage
\appendix
\section{Detailed Definition of Evaluation Metrics}
\label{app:metrics_def}

To ensure the reproducibility and rigor of our empirical analysis, we formally define the eight metrics used across our four evaluative dimensions (as presented in Table~\ref{tab:baseline_performance}).

\paragraph{1. Attack Vulnerability}
These metrics are calculated over the entire evaluation dataset to capture the macroscopic success rate of the adversarial injections.
\begin{itemize}
    \item \textbf{Hijack Rate ($\downarrow$):} The proportion of total evaluation episodes where the agent successfully invokes the attacker-designated tool (e.g., executing a malicious fund transfer or data exfiltration) as a direct result of the injected payload.
    \item \textbf{Utility Preservation ($\uparrow$):} The proportion of episodes where the agent successfully completes the user's original, legitimate task, regardless of whether it was hijacked or not. This measures the operational degradation caused by the injection or the defense mechanism.
\end{itemize}

\paragraph{2. Behavioral Dynamics}
These metrics are computed exclusively on the subset of \textit{hijacked} trajectories to quantify the mechanical rapidity and trajectory deviation caused by the injection.
\begin{itemize}
    \item \textbf{Immediate Compliance Rate (Immed.) ($\downarrow$):} The percentage of hijacked cases where the malicious tool is invoked in the very next action step immediately following the ingestion of the corrupted tool result, indicating a complete absence of multi-step deliberation.
    \item \textbf{Action Divergence Rate (Diverge) ($\downarrow$):} The frequency at which the agent's first action post-injection diverges from the expected first action of a benign, aligned trajectory. It measures the immediate trajectory perturbation caused by the adversarial prompt.
\end{itemize}

\paragraph{3. Linguistic Patterns}
We employ fixed-syntax regex filtering on the agent's \textit{reasoning traces} (the generated ``Thought'' blocks prior to malicious tool execution) to capture semantic compliance. These are computed only on hijacked cases.
\begin{itemize}
    \item \textbf{Resistance Rate (Resist) ($\uparrow$):} The proportion of traces containing explicit refusal, skepticism, or security-alert markers (e.g., matching phrases like \texttt{"I should not"}, \texttt{"suspicious"}, or \texttt{"unauthorized"}). A non-zero rate indicates the model detects the anomaly in its latent reasoning but still mechanically succumbs to the hijack.
    \item \textbf{Semantic Compliance (S-Comp) ($\downarrow$):} The aggregation of the \textit{Immediate} and \textit{Rationalized} thought patterns. In these cases, the reasoning trace contains absolutely no hedging (e.g., no \texttt{"I'm not sure"}) and no resistance. Instead, it either blindly accepts the command or actively generates justifications to frame the malicious action as a legitimate step (e.g., matching phrases that rationalize actions, such as \texttt{"to help the user"} or \texttt{"assist with the task"}). 
\end{itemize}

\paragraph{4. Model Confidence}
These token-level metrics act as proxies for the model's internal certainty during the generation of the hijacked tool-call. 
\begin{itemize}
    \item \textbf{Mean Log-Probability (LogP) ($\uparrow$):} The average log-probability of the generated tokens corresponding to the malicious action. A lower (more negative) LogP compared to benign trajectories indicates latent internal conflict or hesitation during adversarial execution.
    \item \textbf{Mean Entropy ($\downarrow$):} The average Shannon entropy of the model's next-token probability distribution during the hijacked generation step. Abnormally high entropy signifies that the model's predictive distribution is flattened, reflecting uncertainty induced by the conflicting objectives of the system prompt and the injected payload.
\end{itemize}

\section*{LLM Usage Disclosure}
 The authors utilized LLMs solely to enhance the linguistic quality of the manuscript and to assist with data visualization and coding. Every suggestion was thoroughly reviewed by the authors, and they take complete responsibility for all findings and content presented in this paper.

% \section{Appendix}

\end{document}